\documentclass{article}
\usepackage{kr}
\usepackage{multirow}

\usepackage{times}
\usepackage{soul}
\usepackage{url}
\usepackage[hidelinks]{hyperref}
\usepackage[utf8]{inputenc}
\usepackage[small]{caption}
\usepackage{graphicx}
\usepackage{amsmath}
\usepackage{amsthm}
\usepackage{booktabs}
\usepackage{algorithm}
\usepackage{algorithmic}
\usepackage{tikz}
\usepackage{pgfplots}
\pgfplotsset{compat=1.9} 
\usepackage{caption}

\usepackage{array} 
\usepackage{amssymb}

\DeclareMathOperator*{\argmin}{argmin}  
\newcommand{\comment}[1]{}

\title{From Latent to Lucid: Transforming Knowledge Graph Embeddings into Interpretable Structures with KGEPrisma}

\author{%
Christoph Wehner$^{1,2}$\and
Chrysa Iliopoulou$^1$\and
Ute Schmid$^2$\and
Tarek R. Besold$^1$
\affiliations
$^1$Sony AI Barcelona\\
$^2$Cognitive Systems Group, University of Bamberg\\
\emails
\{first.second\}@sony.com
}

\begin{document}

\maketitle

\begin{abstract}
In this paper, we introduce a post-hoc and local explainable AI method tailored for Knowledge Graph Embedding (KGE) models. These models are essential to Knowledge Graph Completion yet criticized for their opaque, black-box nature. Despite their significant success in capturing the semantics of knowledge graphs through high-dimensional latent representations, their inherent complexity poses substantial challenges to explainability. 
While existing methods like Kelpie use resource-intensive perturbation to explain KGE models, our approach directly decodes the latent representations encoded by KGE models, leveraging the smoothness of the embeddings, which follows the principle that similar embeddings reflect similar behaviours within the Knowledge Graph, meaning that nodes are similarly embedded because their graph neighbourhood looks similar. This principle is commonly referred to as smoothness. By identifying symbolic structures, in the form of triples, within the subgraph neighborhoods of similarly embedded entities, our method identifies the statistical regularities on which the models rely and translates these insights into human-understandable symbolic rules and facts. This bridges the gap between the abstract representations of KGE models and their predictive outputs, offering clear, interpretable insights. Key contributions include a novel post-hoc and local explainable AI method for KGE models that provides immediate, faithful explanations without retraining, facilitating real-time application on large-scale knowledge graphs. The method's flexibility enables the generation of rule-based, instance-based, and analogy-based explanations, meeting diverse user needs. Extensive evaluations show the effectiveness of our approach in delivering faithful and well-localized explanations, enhancing the transparency and trustworthiness of KGE models.
\end{abstract}

\section{Introduction}

\begin{figure}[!ht]
\centering
\includegraphics[width=0.3\textwidth]{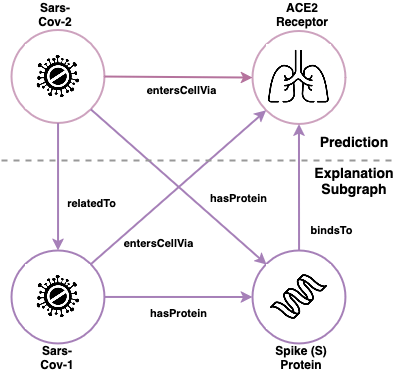}
\caption{KGEPrisma generates explanations for KGE models in the form of subgraphs, uncovering the reasoning of the KGE model and building trust in the model's prediction. For example, the KGE model predicts that Sars-Cov-2 enters the respiratory cell via the ACE2 receptor. The explanation subgraph by KGEPrisma uncover that Sars-Cov-2 has a Spike (S) protein. Furthermore, it shows that Sars-Cov-2 is related to Sars-Cov-1; thus, both are likely to behave similar. Sars-Cov-1 also has a Spike (S) protein, which Sars-Cov-1 uses to bind to the ACE2 receptor, enabling it to enter the respiratory cell. }
\label{fig:example_explanation}
\end{figure}

\begin{figure*}[!ht]
\includegraphics[width=\textwidth]{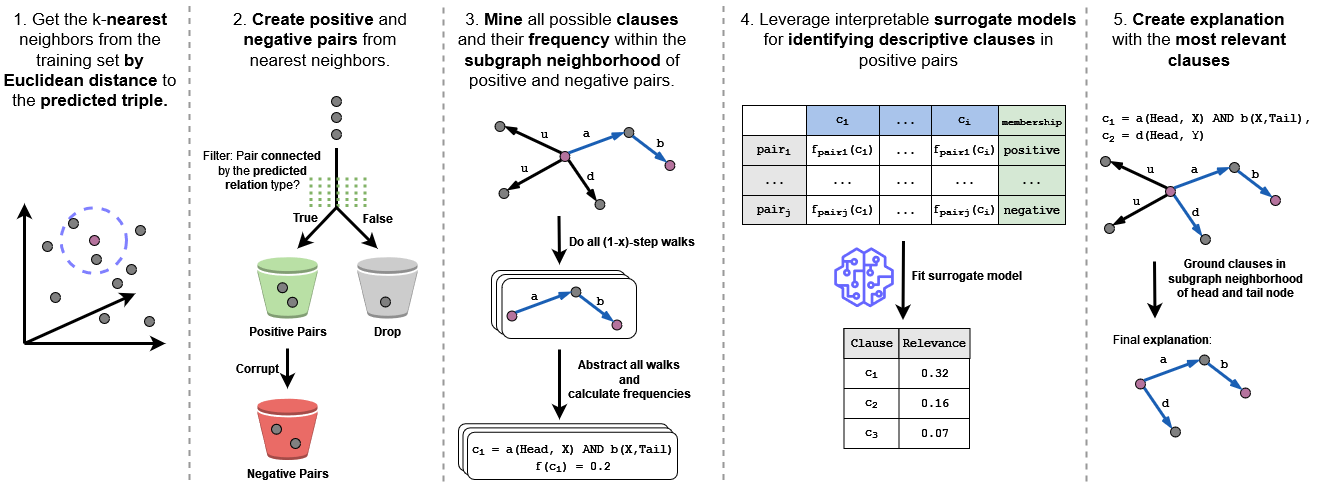}
\caption{KGEPrisma generates explanations of KGE models in five steps. The five steps are discussed in detail in Section~\ref{sec:method}.}
\label{fig:overview}
\end{figure*}

Knowledge Graphs (KG), despite their vast potential for structuring and leveraging information, are notoriously incomplete \cite{ji2022survey,eirich2023manu,BAHR2025100807}. To mitigate this issue, link prediction has emerged as a technique for uncovering previously unknown links within these graphs \cite{hogan2021knowledge}. Knowledge Graph Embedding (KGE) models have become the de facto standard due to their ability to capture the complex relationships and semantics embedded within the graph structure through high-dimensional latent representations \cite{ji2022survey}. However, despite their effectiveness, these models are criticized for their black-box nature \cite{schramm2023comprehensible}, which obscures the underlying mechanisms and rationales behind their predictions \cite{schwalbe_comprehensive_2023}, posing challenges for explainability in critical applications \cite{wehner2022explainable,wehner2023vtc}.

Explainable Artificial Intelligence (XAI) has made significant progress in making the opaque decision-making processes of complex black-box models more transparent \cite{schwalbe_comprehensive_2023}. Despite the development of explainable methods such as LIME \cite{ribeiro2016lime}, SHAP \cite{lundberg2017shap}, Layer-wise Relevance Propagation \cite{Montavon2019lrp}, and Integrated Gradients \cite{sundararajan2017ig}, applying these methods to KGE models presents a non-trivial challenge. These methods traditionally work by attributing parts of the input as relevant or not to the model's output. However, embedding-based link prediction operates differently. It relies on the latent representations of entities and relations in a triple (head, relation, tail) as input to an interaction function to compute a score. This score is then used to create an ordinal ranking of the plausibility of different permutations for the head, relation, or tail \cite{ji2022survey}. In this context, simply assigning relevance to the latent representations of the triple provides minimal insight into the underlying rationale of the prediction. The inherent complexity of these embeddings and the abstract nature of the relations they capture make it difficult to draw clear, interpretable connections between input features and the model's output. 

This work presents the XAI method KGEPrisma. KGEPrisma leverages the principle that KGE models encode a KG's statistical regularities into latent representations, reflecting the KG's structure and interactions.
Central to the method is the smoothness of embeddings, meaning that entities with similar embeddings behave similarly within the KG \cite{bengio_2013}. KGEPrisma decodes these embeddings by identifying distinct symbolic structures in the subgraph neighborhoods of entities with similar embeddings, revealing the model's relied-upon symbolic regularities. These structures can be represented as human-understandable symbolic rules and facts, clarifying the predictive patterns in localized subgraphs (cf.Figure~\ref{fig:example_explanation}). Our evaluations show that the proposed method outperforms state-of-the-art methods regarding faithfulness to the KGE model's decision process and that the explainable evidence is better centered around a region of interest.

This work contributes (1) a novel, local and post-hoc explainable AI method for KGEs. In contrast to others, our method is aligned with the operational mechanics of KGE models, ensuring explanations are faithful to the model's decision-making process, localized around a region of interest and immediate, thereby eliminating the need to retrain the model on occluded training data. This enables real-time, scalable explanations within extensive KGs. (2) Furthermore, our method is versatile, producing explanations in various forms, including rule-based, instance-based, and analogy-based, making it adaptable to diverse user requirements. (3) Through comprehensive evaluations, we demonstrate that our approach performs well compared to existing state-of-the-art methods regarding faithfulness to the model's decision-making process and providing more relevant explanations centered on the user's region of interest.
\section{Preliminaries}

This section briefly introduces Knowledge Graphs, Knowledge Graph Completion (KGC) and Knowledge Graph Embeddings and fixes some notations to be used later. 

\subsection{Knowledge Graph}
A Knowledge Graph is a directed labeled graph $G$ \cite{hogan2021knowledge}, consisting of triples (i.e., facts) $G \subseteq E \times R \times E$ from the entity set $e \in E$ and relation set $r \in R$, allowing the traversal of a triple $(e^{head}, r, e^{tail})$ from a head to a tail entity via a relation. Triples can be expressed as grounded binary predicates $r(e^{head}, e^{tail})$. The relation acts as the binary predicate and the entities as the grounding constants. A KG assigns each entity and relation a symbolic label (e.g., name). KGs are structured according to a semantic schema $s: E \rightarrow C$ \cite{hogan2021knowledge}. This schema categorizes entities into classes $C$ within the KG's domain, facilitating the storage and retrieval of semantically rich, relational data. Nonetheless, the construction of KGs demands substantial expert knowledge, leading to the common issue of incomplete knowledge graphs. 

\subsection{Knowledge Graph Completion}
Knowledge Graph Completion (i.e., Link Prediction) addresses the challenge of inherently incomplete KGs \cite{hogan2021knowledge}. For KGs, there exists a subset of correct but unknown triples $G_{unkown} \subseteq E \times R \times E$  that do not intersect with the existing graph $G$. KGC aims to uncover these missing facts by exploiting the regularities and patterns inherent in the KG, thus deducing the unknown triples. In practice, KGC models are queried with partial triples $(e^{head}, r, ?)$, $(?, r, e^{tail})$, or $(e^{head}, ?, e^{tail})$, seeking to complete these by predicting the missing entity or relation. The model then generates a ranked list of candidates. The higher the rank, the more plausible it is for a candidate to complete the triple \cite{rossi2021survey}.

\subsection{Knowledge Graph Embeddings}
Knowledge Graph Embedding models enable KGC by learning latent space representations $v \in \mathbb{R}^n$ (i.e., embeddings) for entities and relations within a knowledge graph \cite{ali2020benchmarking}, where $n$ is the number of embedding dimensions. An interaction function $i$ assigns a score to the embedding of a triple. 

\begin{equation}
i: E \times R \times E \rightarrow \mathbb{R}
\end{equation}

The score allows the creation of an ordinal ranking. A higher rank indicates a greater plausibility of the triple being true. This scoring mechanism is crucial for optimizing the embeddings to favor existing triples over corrupted ones, ensuring that the embeddings reflect the KG's symbolic regularities. Consequently, embeddings are smooth, meaning that entities exhibiting similar behavior within the graph are represented by similar embeddings \cite{bengio_2013,rossi2021survey,ji2022survey}.
Models such as TransE \cite{bordes2013translating} optimize embeddings by aligning the sum of entity and relation embeddings with the missing entity's embedding. DistMult \cite{yang2015distmult} and ComplEx \cite{blacan2016complex} further refine this approach by implementing a trilinear dot product and extending capabilities to capture non-symmetric relationships. Other models like ConvE \cite{dettmers2018conve} utilize convolutions in the interaction function. Despite the advancements in KGE models, the complexity and abstractness of the embeddings pose significant challenges in establishing clear, interpretable links between input features and model outputs.

\section{Method}
\label{sec:method}

The approach is rooted in the understanding that KGE models encapsulate the symbolic patterns of a KG in smooth latent representations, encoding the graph's topology and the interactions between its entities \cite{bengio_2013}. Thus, entities sharing similar embeddings exhibit comparable behavior within the symbolic structure of the KG. By analyzing the subgraph neighborhoods of these similar entities, symbolic regularities are discovered, in the form of conjunctive clauses \footnote{Clauses are commonly referred to in their disjunctive form, which means that the clause is true whenever at least one of the letters of that form is true. In this paper, clauses are referred to in their conjunctive form. This means that a clause is true when all of the literals that form it are true.} 
, that KGE models depend on. KGEPrisma translates these regularities into symbolic rules, or triples comprehensible to humans, thereby uncovering the rationale behind the models' predictions in local subgraph contexts (cf.Figure~\ref{fig:example_explanation}).
This allows KGEPrisma to post hoc and locally explain the predicted triple $(e^{head}_p, r_p, e^{tail}_p)$, by accessing the training knowledge graph and the embeddings learned by the KGE model. 

The method is build on five steps (cf. Figure~\ref{fig:overview}):
\begin{enumerate}
    \item get k-nearest neighbors in the latent embedding space to the predicted triple (cf. Figure~\ref{fig:overview} Step 1),
    \item create positive and negative entity-pairs from the nearest neighbors (cf. Figure~\ref{fig:overview} Step 2),
    \item mine all possible clauses and their frequency within the subgraph neighbourhood of the pairs (cf. Figure~\ref{fig:overview} Step 3),
    \item identify the most descriptive clauses for positive entity-pairs with the help of a surrogate model (cf. Figure~\ref{fig:overview} Step 4), and
    \item ground the most descriptive clauses to create an explanation (cf. Figure~\ref{fig:overview} Step 5).
\end{enumerate}

In the following section, KGEPrisma is introduced step by step.

\subsection{Step 1: Identifying K-Nearest Neighbors} \label{subsec:knn-search}
In the initial step of the post hoc explainability method, the embedding $v_{predicted}$ of a given predicted triple is taken as input. The creation of triple embeddings is KGE model-dependent. Appendix~\ref{app:triple_embeddings} provides examples of KGE models and their methods for obtaining triple embeddings. In its simplest form, $v_{predicted} = [v^{head};v^{r};v^{tail}]$ is a concatenation of the entity and relation embeddings.

The k-nearest neighbor embeddings ${v_1,v_2,...,v_k}$ are then retrieved from the set of all KGE model training triple embeddings $V_{train}$, based on the Euclidean distance (cf. Figure~\ref{fig:overview} Step 1). The Euclidean distance is used as it demonstrates robust performance in finding similarity-based explanations in previous work conducted on image data \cite{hanawa2021evaluation} \footnote{The cosine distance was used in an ablation study; however, no meaningful impact on the method was observed.}. The retrieval is described by the equation:
\begin{equation}
    kNN(v_{predicted})= \argmin_{v \in V_{train}}{}_k ||v_{predicted} - v||_2
\end{equation}

In this equation, $\argmin_{}{}_k$ identifies the $k$ embeddings $v$ that yield the smallest Euclidean distances to $v_{predicted}$, thus isolating the embeddings in the latent space that are most likely to exhibit symbolic regularities in common with the predicted triple. Thus, KGEPrisma considers the local decision surface of the KGE model to explain a triple, enabling an efficient computation and coupling the explanation closely to the model behaviour. 
This step guarantees that the explanation generated in downstream steps reflects the internal mechanics of the KGE model by localizing the explanation around the training instances that the model learned to see and treat similarly.
The embeddings are then mapped back to their symbolic triple representations, the relationship symbol is dropped and the entity-pairs are stored in $N = ((e^{head}_1, e^{tail}_1), (e^{head}_2, e^{tail}_2), ...,(e^{head}_k, e^{tail}_k))$. In the next step, positive and negative pairs are created with the help of the pairs in $N$. 

\subsection{Step 2: Create Positive and Negative Entity-Pairs}

Step two constructs positive and negative entity pairs (cf. Figure~\ref{fig:overview} Step 2). A nearest neighbor pair $(e_i, e_j) \in N$ belongs to the positive set $P^{+}$ if $(e_i, r_{p}, e_j)$ is a fact in $G$, where $r_p$ is the relation type of the predicted triple we want to explain. Conversely, a pair $(e_k, e_l)$ is in the negative set $P^{-}$ if $(e_k, r_{p}, e_l)$ does not exist in $G$, representing a corrupted version of a positive pair.

\begin{align}
    P^{+}=\{(e_i,e_j)\in N \vert (e_i, r_p, e_j) \in G\}  \nonumber\\ 
    P^{-}=\{ (e_k,e_l) \vert (e_k = e_i \veebar e_l = e_j)  \\ \land (e_k,r_p,e_l)\notin G \}, \nonumber \\
    with \  (e_i,e_j) \in P^{+} \land e\in E  \nonumber
\end{align}

The process results in two sets, $P^{+}$ containing pairs that are connected by the predicted relation type and have a similar latent representation to the predicted triple, and $P^{-}$, which includes pairs that serve as corrupted versions of the positive pairs. It is important to emphasize that $P^{+}$ holds the triples from which the KGE model learned its behavior. In practice, one corrupted pair for every positive pair in $P^{+}$ is sampled by randomly switching a positive pair's head or tail entity with a random entity from the KG. This procedure is similar to the stochastic local closed-world assumption applied while training KGE models \cite{ali2020benchmarking}. 

\subsection{Step 3: Mining Clauses Frequency} \label{subsec:mining_clauses}

In the third step, the method abstracts the subgraph neighborhoods of entity pairs into conjunctive clauses and computes their frequencies (cf. Figure~\ref{fig:overview} Step 3). This process aims to identify symbolic regularities, in the form of clauses, that the KGE model implicitly relies on for its predictions.

\begin{algorithm}[H] 
\caption{Clause Mining and Frequency Calculation} \label{alg:clause_mining}
\textbf{Input}: Positive pairs $P^+$, negative pairs $P^-$, knowledge graph $G$ \\ 
\textbf{Parameter}: Maximum walk length $x$ \\ 
\textbf{Output}: Dictionary $D$ mapping each pair to its unique clauses and frequencies \begin{algorithmic}[1] 
\FOR{each pair $(e_{\text{head}}, e_{\text{tail}}) \in P^+ \cup P^-$} 
\STATE Initialize an empty multiset $S_{(e_{\text{head}}, e_{\text{tail}})}$ to  \\ store clauses
\FOR{each walk $w$ in $G$ of length $1$ to $x$, starting or ending at $e_{\text{head}}$ or $e_{\text{tail}}$, without including them as intermediate nodes} 
\STATE Apply schema mapping $s: E \rightarrow C \cup \{ Head, Tail \}$ to abstract entities in $w$ 
\STATE Obtain clause $c$ from the abstracted walk 
\STATE Add $c$ to $S_{(e_{\text{head}}, e_{\text{tail}})}$ 
\IF{length of $w$ equals $1$} 
\STATE Abstract only the head and tail entities in $w$ to $\text{Head}$ and $\text{Tail}$ 
\STATE Obtain clause $c'$ from the partially abstracted walk 
\STATE Add $c'$ to $S_{(e_{\text{head}}, e_{\text{tail}})}$ 
\ENDIF 
\ENDFOR 
\FOR{each unique clause $c$ in $S_{(e_{\text{head}}, e_{\text{tail}})}$} 
\STATE Compute frequency $f_c = \frac{|c \in S_{(e^{head}, e^{tail})}|}{|S_{(e^{head}, e^{tail})}|}$ \STATE Store $(c, f_c)$ in $D_{(e_{\text{head}}, e_{\text{tail}})}$ 
\ENDFOR 
\ENDFOR 
\end{algorithmic} 
\end{algorithm}

For each entity pair $(e_{\text{head}}, e_{\text{tail}})$ in the combined set $P = P^+ \cup P^-$, walks of length from $1$ to $x$ are constructed in the knowledge graph $G$. These walks start or end at either $e_{\text{head}}$ or $e_{\text{tail}}$, but do not include them as intermediate nodes. Each walk $w$ represents a sequence of entities and relations anchored in $e_{\text{head}}$ or $e_{\text{tail}}$.

To abstract the walks into clauses, a schema mapping $s: E \rightarrow C \cup \{\text{Head}, \text{Tail}\}$ is applied, where $E$ is the set of entities, $C$ is the set of classes, and $\text{Head}$ and $\text{Tail}$ are special classes assigned to $e_{\text{head}}$ and $e_{\text{tail}}$, respectively. This mapping replaces each entity in the walk with its class, creating an abstract representation of the walk that preserves structural patterns while generalizing over specific entities.

Each abstracted walk corresponds to a conjunctive clause $c$, which can be thought of as a logical expression capturing the relationships between classes in the neighborhood of the entity pair. By including the special classes $\text{Head}$ and $\text{Tail}$, the abstraction maintains anchors to the original entities, enabling later grounding.


The method thus captures following conjunctive clause types:
\begin{align}
    r_1(A, W) \land r_2(X, Y) \land ... \land r_m(Y, B)\\
    r_1(A, W) \land r_2(X, Y) \land ... \land r_m(Y, Z)\\
    r_1(X, Y) \land r_2(Y, Z) \land ... \land r_m(Z, A), 
\end{align}
where $A, B \in \{Head,Tail\}$, the variable $m \leq x$ is the actual step length, and $W$, $X$, $Y$, $Z \in C$ are classes.

For each unique clause $c$ obtained from the walks associated with an entity pair, its frequency $f_c$ is calculated. The frequency is the proportion of times the clause appears in the set of all clauses $S_{(e_{\text{head}}, e_{\text{tail}})}$ for that pair. This frequency reflects the relative significance of the clause in the subgraph neighborhood of the entity pair.

The tuple $(c, f_c)$ is stored in the dictionary $D_{(e_{\text{head}}, e_{\text{tail}})}$ for each pair, resulting in a mapping of each entity pair to its unique clauses and their frequencies. Algorithm~\ref{alg:clause_mining} details this procedure.

This process ensures that the frequencies of clauses are mapped for each entity pair, which are then used in the following step to identify the symbolic regularities that the KGE model relies on for its predictions.

\subsection{Step 4: Leveraging Surrogate Models for Identifying Descriptive Clauses in Positive Entity Pairings} \label{subsec:surrogate_models}

The dictionary $D$ establishes a classic tabular machine learning setup, wherein instances are represented by entity pairs, features by clauses, and values by the frequency of the clauses. The labels are categorized as positive or negative based on the entity pair's membership of $P^+$ or $P^-$ (cf. Figure~\ref{fig:overview} Step 4). The objective is to identify which feature (clause) contributes the most to classifying an entity pair as positive. This is achieved by utilizing surrogate models, which allows extracting the feature importances to interpret the complex relationships within the data.

The goal is thus to assign each clause a score by which it is ranked in accordance with its relevance for classifying an entity pair as positive or negative. 

For KGEPrisma mean decrease in impurity \cite{nembrini2018revival}, K-Lasso \cite{ribeiro2016lime}, and HSIC-Lasso \cite{yamada2014high} are studied and compared to identify feature importance.

The \textbf{Mean Decrease in Impurity} (MDI) quantifies each clause's role in classifying positive or negative samples in $D$ through an ensemble of decision trees \cite{nembrini2018revival}. This process entails iterative data splitting based on clauses that maximize impurity reduction, employing the Gini impurity \cite{nembrini2018revival} as a measure of this reduction. The Gini impurity for a dataset $d \subseteq D$ is defined as:

\begin{equation}
    Gini(d) = 1 - \sum_ip(i|t)^2
\end{equation}

where $p(i|t)$ denotes the proportion of class $i \in \{positive, negative\}$ at tree-node (dataset) $d \subseteq D$, adjusted by weights $\alpha$ reflecting the Euclidean distance of a pair embedding $v_i$ to the predicted pairs's embedding $v_p$. The weight is defined as an exponential kernel $\alpha(v_i)= \exp (-||(v_p, v_i)||_2^2/\sigma^2)$ with kernel width $\sigma$. This assigns a higher impact to pairs that are perceived by the model as similar. The Gini impurity evaluates the likelihood of mislabeling an element if randomly assigned based on the subset's label distribution, serving as a statistical regularity indicator in $D$.
Most relevant features reduce the Gini impurity of a dataset by the most over all nodes within the tree. 

The impurity reduction ($\Delta Gini$) \cite{nembrini2018revival} from splitting at tree-node $d$ on clause $c$, yielding "positive" ($L$) and "negative" ($R$) child nodes, is given by:

\begin{equation}
    \Delta Gini(d, c)=Gini(d)-(\frac{N_L}{N_d}Gini(L)+\frac{N_R}{N_d}Gini(R))
\end{equation}

Here, $N_d$, $N_L$, and $N_R$ represent the weighted counts of samples at the parent node and in each child node, respectively.

The MDI for a clause across the ensemble is the impurity reductions' mean, weighted by the samples reaching the nodes where the feature splits the data:
\begin{equation}
    MDI(c) = 1 - \frac{\gamma_c}{N} \sum_{d_c\in D}\Delta Gini(d_c, c)  N_d
\end{equation}

where $d_c$ is a node split on clause $c$, and $N_d$ is the total sample count in $d_c$. A weighted frequency co-factor $\gamma$ is applied to the MDI of a clause. It is defined as: 
\begin{equation} \label{equ:frequency_cofactor}
\gamma(c) = 
\begin{cases}
    1  & \text{if } \sum_{f_c \in D_+} f_c \geq \sum_{f_c \in D_-} f \\
    -1  & \text{otherwise}
\end{cases}
\end{equation}
This allows to weight in if a clause is more frequent in positive instances $D_+$ or negative instances $D_+$ of the dictionary. 

MDI thereby assesses clause importance, identifying those crucial for positive pair classifications within $D$, revealing key statistical patterns in similar sub-graph neighborhoods. 
Nonetheless, MDI may favor features with higher cardinality, such as those capturing multi-hop regularities, over binary property relations, due to inherent biases of MDI toward features with broader variation \cite{nembrini2018revival}.

The \textbf{K-Lasso} method uses a linear model, specifically ridge regression \cite{arthur2000ridge}, to weight each clause contribution in the classification task within the dictionary $D$. The method learns a weight for every feature (clause), employing linear least squares with Euclidean regularization to optimize the model \cite{ribeiro2016lime}.
The objective function minimized by this model is formalized as:

\begin{equation}
    min(\sum_{(e_i, e_j)\in D}\alpha_{(e_i, e_j)}(y_{(e_i, e_j)} - C_{(e_i, e_j)}^Tw) +\beta||w||_2^2)
\end{equation}

Here, $y \in \{1,-1\}$ is the label (positive, negative) of the entity pair $(e_i, e_j) \in D$, $C$ is the feature vector holding the frequencies of all clauses of an entity pair, and $w$ is the vector of weights corresponding to the clauses. The kernel $\alpha (v_{(e_i, e_j)})= \exp (||(v_p, v_{(e_i, e_j)})||_2^2/\sigma^2)$ with kernel width $\sigma$ scales the impact of pairs that are perceived by the model as similar, allowing for differential emphasis on instances that are perceived by the model as closer to the predicted triple.
The parameter $\beta$ is for the Euclidean regularization, penalizing the sum of squared weights to prevent overfitting.
After fitting the surrogate model, the learned weights $w$ for each feature provide a direct measure of feature importance. These weights reflect the contribution of each clause to the prediction task, with larger absolute values indicating greater importance. This enables the identification of the most relevant clauses that contribute to classifying entity pairs as positive or negative, providing insights on the underlying statistical regularities captured by the KGE models.

Compared to MDI, K-Lasso is not biased towards features with high cardinality. However, it permits only linear feature selection, which may not capture complex relationships between features in certain datasets effectively \cite{yamada2014high}.

The \textbf{Hilbert-Schmidt Independence Criterion Lasso} (HSIC-Lasso) is a supervised nonlinear feature selection methodology aimed at identifying a subset of input features relevant to predicting output values. As an extension of the standard Lasso, HSIC-Lasso incorporates a feature-wise kernelized Lasso to capture nonlinear dependencies between inputs and outputs. This enables it to identify non-redundant features with a significant statistical dependence on the output values \cite{yamada2014high,huang2023graphlime}.

The optimization problem of HSIC-Lasso is formalized as follows \cite{yamada2014high}:
\begin{align}
\min_{w_1,\ldots,w_d} \frac{1}{2} || \mathbf{L} - \sum_{k} w_k \mathbf{\tilde{K}}^{(k)} ||_F^2 + \lambda \sum_{k} |w_k| \nonumber \\ \quad , with \quad w_1,\ldots,w_d \geq 0
\end{align}
where $\left| \cdot \right|_F$ denotes the Frobenius norm, $\mathbf{\tilde{K}}^{(k)}$ represents the centered Gram matrix for the $k$-th feature, and $\mathbf{L}$ is the centered Gram matrix for the output $y$.

After training the surrogate model, coefficients ($w$) are obtained, which, when multiplied by the frequency co-factor $\gamma$ (cf. Equation \ref{equ:frequency_cofactor}), identify clauses predominantly associated with the sub-graph neighborhood of positive entity pairs. The coefficients reflect how relevant each clause is to the prediction.

\subsection{Step 5: Generating Explanations from the Most Descriptive Clauses}

After obtaining the most descriptive clauses from the surrogate model, they are used to generate explanations of the KGE model's prediction (cf. Figure~\ref{fig:overview} Step 5). The approach allows generating three explanation types, each catering to different aspects of user needs: rule-based, instance-based, and analogy-based. It is important to note that all explanation types are based on the same clauses. The difference is in their grounding triples, which gives the user different perspectives on why the model believes the predicted triple is true and thus increases the trust in the prediction. Table~\ref{tab:explanation_types} provides an example of that. 

\begin{table*}[!ht]
\centering 
\caption{Comparison of explanation types generated from examplatory most descriptive clauses, given the predicted triple $(Alice, knows, Bob)$ and its closest positive neighbour $(Carol, Dave)$. For simplicity, the table shows only the two most relevant clauses identified by KGEPrisma.}
\resizebox{0.8\textwidth}{!}{%
\begin{tabular}{c||p{4.5cm}|p{4.5cm}|m{1cm}} 
& \textbf{1st Clause} & \textbf{2nd Clause} & ... \\
\hline
\hline
\textbf{Clause} & $knows(Head,Person)$\newline $\land \ works\_with(Person,Tail)$ & $knows(Head,Person)$ $\land \  sibling\_of(Person,Tail)$ & ... \\
\hline
\textbf{Relevance} & 0.54 & 0.31 & ... \\
\hline
\textbf{Rule-Based Explanation} & $knows(Alice,Person)$ \newline $\land \  works\_with(Person,Bob)$ \newline $\rightarrow \ knows(Alice,Bob)$ &  $knows(Alice,Person)$ \newline $\land \  sibling\_of(Person,Bob)$ \newline $\rightarrow \ knows(Alice,Bob)$ & ... \\
\hline
\textbf{Instance-Based Explanation} & $knows(Alice,Tom)$ \newline $\land \  works\_with(Tom,Bob)$ & $knows(Alice,Pedro)$ \newline $ \land \ sibling\_of(Pedro,Bob)$ & ... \\
\hline
\textbf{Analogy-Based Explanation} & $knows(Carol,Anja)$ \newline $\land \  works\_with(Anja,Dave)$ & $knows(Carol,Jan)$ \newline $\land \  sibling\_of(Jan,Dave)$ & ... \\
\hline
\end{tabular}
}
\label{tab:explanation_types}
\end{table*}

\textbf{Rule-based} explanations are derived by appending an implication to the most relevant clauses, thus forming a set of symbolic rules. These rules express the symbolic regularities that the KGE model has learned to predict a missing link. For instance, if the most descriptive clause extracted for the predicted triple $(Alice,knows,Bob)$ is $knows(Head,Person)\land works\_with(Person,Tail)$, the rule is formulated as $knows(Alice,Person)\land works\_with(Person,Bob)\rightarrow knows(Alice,Bob)$. This implies that if $Alice$ $knows$ someone of the class $Person$, and this $Person$ $works\_with$ $Bob$, the triple $(Alice,knows,Bob)$ is predicted. Rule-based explanations provide the user with a broader justification of why the predicted triple is believed to be true by showing a pattern that is dominant in the subgraphs around similar embedded training triple instances. 

\textbf{Instance-based} explanations are generated by grounding the most relevant clauses in the knowledge graph. Grounding, in logical terms, means replacing the variables in a clause with specific constants from the domain, thus instantiating the clause. For example, if $knows(Head,Person)\land works\_with(Person,Tail)$ is a clause and the predicted triple is $(Alice,knows,Bob)$, the grounding would be $knows(Alice,Tom)\land works\_with(Tom,Bob)$. This means that $Alice$ $knows$ $Tom$, and $Tom$ $works\_ with$ $Bob$; thus, $Alice$ $knows$ $Bob$. This type of explanation provides the concrete triples that led the model to predict the $knows$-relation between $Alice$ and $Bob$.

\textbf{Analogy-based} explanations focus on how the model behaves in similar situations by grounding the literals with the head and tail of the pair from $P^+$ that is closest in terms of Euclidean distance to the predicted pair. This approach demonstrates the model's behavior on similar instances, for which the prediction is confirmed to be true by the training facts. For example, if the nearest pair to $(Alice,knows,Bob)$ in $P^+$ is $(Carol,Dave)$, and $knows(Head,Person)\land works\_with(Person,Tail)$ is a clause, the grounding would be $knows(Carol,Anja)\land works\_with(Anja,Dave)$. This means that $Alice$ $knows$ $Tom$ because first of all $Carol$ and $Dave$ are similar to $Alice$ and $Tom$. And for $Carol$ and $Dave$, it is a fact that they $know$ each other because $Coral$ $knows$ $Anja$, and $Anja$ $works\_with$ $Dave$. Thus, because this pattern also works by analogy for $Alice$ and $Tom$, $Alice$ has a high chance of knowing $Tom$. This shows an analogous situation where the model applied similar decision-making.

The resultant triples are then presented to the user. This allows the user to uncover the hidden symbolic regularities that the KGE model has learned and utilized to predict the missing link. Such explanations not only enhance the transparency of the model but also increase the user’s trust by making the model's predictions interpretable and verifiable.

\section{Evaluation}

\begin{table*}[ht]
\centering
\caption{Results for FB15k-237. The results are the mean and variance of the MRR and Hits@1 over ten runs; the lower the MRR and Hits@1, the better. The best results are bold.}
\resizebox{\textwidth}{!}{%
\begin{tabular}{ c | c c | c c | c c }
 & \multicolumn{2}{c | }{TransE} & \multicolumn{2}{c|}{DistMult} & \multicolumn{2}{c}{ConvE} \\
\hline
 & MRR & Hit@1 & MRR & Hit@1 & MRR & Hit@1 \\
 \hline
Retraining & 0.97 $\pm$ 0.03 & 0.94 $\pm$ 0.06 & 0.84 $\pm$ 0.12 & 0.78 $\pm$ 0.16 & 0.84 $\pm$ 0.07 & 0.71 $\pm$ 0.00 \\
\hline
Global Random & 0.96 $\pm$ 0.03 & 0.93 $\pm$ 0.05 & 0.82 $\pm$ 0.15 & 0.69 $\pm$ 0.27 & 0.76 $\pm$ 0.22 & 0.54 $\pm$ 0.43 \\ 
Local Random & 0.96 $\pm$ 0.03 & 0.93 $\pm$ 0.04 & 0.70 $\pm$ 0.24 & 0.51 $\pm$ 0.36 & 0.69 $\pm$ 0.38 & 0.64 $\pm$ 0.43 \\
\hline
AnyBURLExplainer & 0.95 $\pm$ 0.02 & 0.92 $\pm$ 0.05 & 0.41 $\pm$ 0.35 & 0.25 $\pm$ 0.26 & 0.40 $\pm$ 0.25 & 0.16 $\pm$ 0.29 \\
Data Poisoning & 0.96 $\pm$ 0.03 & 0.92 $\pm$ 0.05 & 0.40 $\pm$ 0.37 & 0.26 $\pm$ 0.29 & 0.41 $\pm$ 0.31 & 0.22 $\pm$ 0.36 \\
Kelpie & \textbf{0.94} $\pm$ 0.03 & \textbf{0.90} $\pm$ 0.06 & 0.37 $\pm$ 0.38 & 0.27 $\pm$ 0.33 & 0.49 $\pm$ 0.32 & 0.30 $\pm$ 0.36 \\
\hline
KGEPrisma - K-Lasso & 0.95 $\pm$ 0.04 & 0.92 $\pm$ 0.06 & 0.34 $\pm$ 0.08 & \textbf{0.00} $\pm$ 0.00 & 0.41 $\pm$ 0.37 & 0.28 $\pm$ 0.41 \\
KGEPrisma - HSIC-Lasso & 0.95 $\pm$ 0.02 & 0.92 $\pm$ 0.04 & \textbf{0.00} $\pm$ 0.00 & \textbf{0.00} $\pm$ 0.00 & \textbf{0.11} $\pm$ 0.09 & \textbf{0.00} $\pm$ 0.00 \\
KGEPrisma - MDI & 0.95 $\pm$ 0.03 & 0.91 $\pm$ 0.05 & \textbf{0.00} $\pm$ 0.00 & \textbf{0.00} $\pm$ 0.00 & 0.32 $\pm$ 0.32 & 0.19 $\pm$ 0.37 \\
\end{tabular}
}
\label{tab:results_fb15k237}
\end{table*}

The evaluation section outlines the protocol to assess the faithfulness \cite{hedstrom2023quantus} of KGEPrisma to KGE model behavior, comparing it against a retraining baseline, a local random baseline, a global random baseline, and the state-of-the-art methods AnyBURLExplainer \cite{ijcai2022p391}, Data Poisoning \cite{zhang2019data}, and Kelpie \cite{rossi2022kelpie}, using the Kinship \cite{kemp2006kinship}, WN18RR \cite{bordes2013translating}, and FB15k-237 \cite{toutanova2015fb15k237} KGs. 
Additionally, a runtime evaluation of KGEPrisma and a qualitative evaluation of the three explanation types (instance-based, rule-based, analogy-based) are presented in this section. 

\subsection{Evaluation Setting}
The evaluation involves the three benchmark KGs FB15k-237, WN18RR, and Kinship designed for evaluating KGE models.
FB15k-237 \cite{toutanova2015fb15k237} is derived from FB15k to address the challenge of inverse relation test leakage. FB15k is built from the Freebase repository, which covers a wide range of domains such as music, films, locations, books, and people. FB15k's splits were problematic because many triples were inverses of each other, leading to leakage between training and testing datasets. FB15k-237 improves upon this by excluding inverse relations, thus containing 310,079 triples, 14,505 entities, and 237 relation types. The following evaluations used the standard train, test, and validation split. 

WN18RR \cite{bordes2013translating} was developed as an improvement over the WN18 KG, a WordNet semantic network subset. WordNet describes semantic and lexical connections between terms, for instance, hyponyms, hypernyms, and antonyms. WN18RR was created to eliminate the issue of inverse relation leakage present in WN18. It includes 93,003 triples, maintaining the same set of 40,943 entities, but reduces the number of relation types to 11 from the original 18 in WN18. The following evaluations used the standard train, test, and validation split. 

The Kinships \cite{kemp2006kinship} KG maps the kinship relationships within the Alyawarra tribe from Australia. It comprises 10,686 triples, 104 entities, and 26 types of relations. The entities represent tribe members, and the relations are defined by specific kinship terms like Adiadya and Umbaidya, reflecting the social structure and rules of familial ties within the tribe. Kinship has a high number of inverse relations, making it suboptimal for assessing the performance of KGE models. However, this enables the evaluation of whether the explainability method can capture the KGE model's dependency on inverse relations and its ability to pinpoint explanations in such regions of interest (e.g., $brotherOf(Tom,Hans) \rightarrow brotherOf(Hans,Tom)$). The following evaluations used the standard train, test, and validation split. 

For the evaluation the three KGE models TransE, DistMult, ConvE are trained on the KG's. The three models were chosen as examples of KGE models with diverse interaction functions. 

TransE \cite{bordes2013translating} is an energy-based model that produces embeddings by interpreting relationships as translations in a low-dimensional space (cf. Appendix~\ref{app:TransE}). If a relationship holds, the embedding of the tail entity should be proximal to the summation of the head entity embedding and the relation embedding.

DistMult \cite{yang2015distmult} simplifies the RESCAL model \cite{nickel2011rescal} by representing relationships with diagonal matrices instead of full matrices (cf. Appendix~\ref{app:DistMult}). This reduction in computational complexity comes at the expense of expressive power, as DistMult cannot model anti-symmetric relations.

ConvE \cite{dettmers2018conve} utilizes a convolutional architecture, which includes a single convolution layer, a projection layer, and an inner product layer (cf. Appendix~\ref{app:ConvE}). It is parameter-efficient and effective in modeling nodes with high in-degree, which is common in complex knowledge graphs.

These models were configured based on a comprehensive hyperparameter optimization study conducted by \cite{ali2020benchmarking}. Details about the embeddings they produce are provided in the Appendix~\ref{app:triple_embeddings}. The PyKEEN library \cite{ali2020benchmarking} is used for implementing the KGE models. 

The evaluation assesses KGEPrisma using the three different surrogate model configurations: MDI, K-Lasso, and HSIC-Lasso (cf. Section~\ref{subsec:surrogate_models}), all utilizing $k=40$ for nearest neighbor search in step 1 (cf. Section~\ref{subsec:knn-search}). Additionally, the maximal clause length is set to $t=2$ for FB15k-237, to $t=3$ for WN18RR, and to $t=1$ for Kinship in step 3 (cf. Section~\ref{subsec:mining_clauses}). This hyperparameter configuration performed best in an ablation study. 

The evaluation of KGEPrisma is conducted against a simple retraining pipeline, two random baselines, AnyBURLExplainer \cite{ijcai2022p391}, Data Poisoning \cite{zhang2019data}, and Kelpie \cite{rossi2022kelpie}. 

The retraining baseline retrains the KGE model without any changes to the training set. However, a different random seed may result in different outcomes. 

The global random baseline selects a random path from the training KG as the explanation, which poses the risk of choosing irrelevant paths in large KGs. The local random baseline, however, selects a random path either starting or ending at the predicted head or tail node, utilizing triples close to the predicted triple, thereby impacting the prediction more significantly and creating a stronger baseline. Both baselines are adjusted to a path length of $2$ for FB15k-237, of $t=3$ for WN18RR, and of $t=1$ for Kinship to align with KGEPrisma. 

The state-of-the-art method by \cite{ijcai2022p391} is used to compare the KGEPrisma against existing literature. The method by \cite{ijcai2022p391} is called AnyBURLExplainer in the following. AnyBURLExplainer utilizes AnyBURL \cite{meilicke2019anyburl} to learn rules from a training knowledge graph, which are then used to attack and explain predicted triples. The evaluation of AnyBURLExplainer is based on an implementation provided by its developers \footnote{AnyBURLExplainer implementation: https://web.informatik.uni-mannheim.de/AnyBURL/}, limiting the rule length to 2 and the rule learning time to 100 seconds.

Data Poisoning \cite{zhang2019data} is an adversarial attack method used for KGE models. The aim of this method is to manipulate the training set by adding or removing specific triples, which in turn alter the plausibility score assigned to a target triple by the KGE model. This manipulation is accomplished by shifting the embedding of the target triple based on the gradient of the KGE model's scoring function, with the goal of identifying adversarial triples. While data poisoning primarily serves as an adversarial attack method, the adversarial triples it generates can also be utilized as explanation triples. The implementation provided by \cite{rossi2022kelpie} is used for this evaluation \footnotemark{}.

Kelpie \cite{rossi2022kelpie} is an explainability framework designed for KGE models. It post-hoc identifies subsets of training facts that are either necessary or sufficient to explain specific predictions. Kelpie relies on heuristics such as holistic mimics and prefiltering to reduce the search space for explanations. 
A necessary explanation is the smallest set of facts such that removing these facts from the training set and retraining the model results in the prediction being false.
A sufficient explanation is the smallest set of facts such that their addition to a set of entities enables the model to predict a specific tail of a query. The notion of faithfulness aligns with necessary explanations. Thus, necessary explanations produced by Kelpie are used for evaluation. The implementation of Kelpie provided by its developers is used for this evaluation \footnotemark[\value{footnote}].

\footnotetext{Data Poisoning and Kelpie implementation: https://github.com/AndRossi/Kelpie/tree/master}

KGEPrisma, AnyBURLExplainer, and Kelpie provide multiple triples to explain one predicted triple. This allows for an evaluation protocol that captures the full expressiveness of the post-hoc XAI methods.  

\subsection{Faithfulness Evaluation Protocol}

The evaluation protocol is designed to fully utilize an XAI method's expressivity, measuring its faithfulness by incorporating all triples that are part of the explanation. This approach ensures a comprehensive assessment of the method's performance.

\textbf{Faithfulness} \cite{hedstrom2023quantus}, in this context, measures the degree to which explanations mirror the predictive behavior of the KGE model, asserting that more crucial features have a stronger influence on model decisions. This relationship is traditionally verified in XAI through input perturbation, which means removing most relevant features and observing any decline in model performance \cite{hedstrom2023quantus}. However, input perturbation is not feasible for KGE models.

Instead, this paper proposes a \textbf{protocol} that includes training a KGE model on the complete training set $D$ and selecting 30 optimal-performing triples $P$ from the validation set, achieving a Hits@1 \cite{ali2020benchmarking} and MRR \cite{ali2020benchmarking} of 1 on $P$. An explanation $E$ is generated for each triple in $P$, incorporating all grounding triples for the explanation rule or clause. These triples are then removed from $D$ to form a new training set $D' = D \setminus E$, on which the model is retrained from scratch. The performance of $P$ on this retrained model is assessed by Hits@1 and MRR, with values closer to 0 indicating a higher model deterioration and thus a more faithful explanation. As pointed out by \cite{ijcai2022p391}, it is crucial that the filter set used to calculate Hits@1 and MRR remains unaltered by $E$ to avoid artificially deflating model performance and to genuinely measure the impact of the explanation.

\subsection{Faithfulness Evaluation Results and Discussion}

This section presents the evaluation results for the faithfulness evaluation protocol.

The results for \textbf{FB15k-237}, as detailed in Table \ref{tab:results_fb15k237}, indicate that KGEPrisma consistently outperforms the other XAI methods across all KGE models. 

For TransE, all methods demonstrate comparably poor performance, not meaningfully surpassing random baselines. The triples that TransE predicts on FB15k-237 are primarily self-loop relations, such as $/location/hud\_county\_place/place$, which connects the entity $Kansas\ City$ to itself, or the self-loop relation $/education/educational\_institution/campuses$, linking $Virginia\ Tech$ to itself. These self-loop relations are self-entailed and do not depend on patterns within the subgraph neighborhood. The model simply needs to remember them.
This suggests that TransE's simplistic interaction function fails to effectively capture logical rules or patterns in FB15k-237. Instead, it relies on reconstructing relations from memory, which primarily works for self-loop relations and is hard to capture via triple-based explanations. 

In the case of DistMult, KGEPrisma demonstrates the best performance compared to the other methods. Particularly noteworthy is its low variance. KGEPrisma with K-Lasso performs on par with other methods. However, when paired with HSIC-Lasso and MDI surrogate, KGEPrisma performs exceptionally well, deteriorating DistMult's MRR and Hits@1 to zero after retraining. This success can be attributed to KGEPrisma’s ability to consistently identify better explanations.

\begin{figure}[!ht]
\centering
\includegraphics[width=0.5\textwidth]{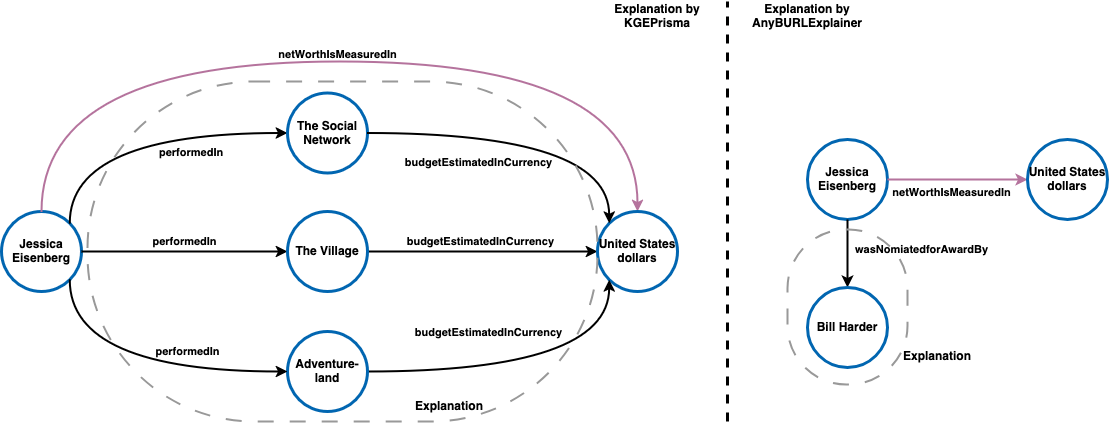}
\caption{
DistMult in FB15k-237 predicts that Jessica Eisenberg's net worth is measured in US dollars. KGEPrisma explains this by pointing to the movies Jessica Eisberg performed in and their budget currency, US dollars, which is a sensible explanation. AnyBURLExplainer points to the fact that Jessica Eisberg was nominated alongside an award with Bill Hader as an explanation for Jessica Eisenberg's net worth being measured in US dollars. This explanation is not sensible.}
\label{fig:jessica}
\end{figure}

For instance, DistMult predicts that $Jessica\ Eisenberg's$ net worth is measured in (i.e. relation: $/base/schemastaging/person\_extra/net\_worth.$ $/measurement\_unit/dated\_money\_value/currency$) $United\ States\ dollars$ (cf. Figure~\ref{fig:jessica}), which is true. KGEPrisma explains this prediction by stating that $Jessica\ Eisenberg$ performed in the films (i.e. relation: $/film/actor/film./film/performance/film$) $The\ Social\ Network$, $Adventureland$, and $The\ Village$. Since the budgets for these films (i.e. relation: $/film/film/estimated\_budget./measurement\_unit$ $/dated\_money\_value/currency$) are measured in $United\ States\ dollars$, it is quite plausible that $Jessica\ Eisenberg's$ net worth is also measured in $United\ States\ dollars$.
In contrast, other XAI methods provide seemingly unrelated explanations. For example, AnyBURLExplainer explains the same predicted triple by stating that $Jessica\ Eisenberg$ was nominated (i.e. relation: $/award/award\_nominee/award\_nominations./award$ $/award\_nomination/award\_nominee$) alongside $Bill\ Hader$ for an award, which appears irrelevant to the predicted triple. This disconnection in the predicted triple versus explanation is further reflected in their high variability in faithfulness scores.  

In the case of ConvE on FB15k-237, the results are more mixed. Here, KGEPrisma with HSIC-Lasso outperforms the other methods by a significant margin. The other XAI methods, besides KGEPrisma with HSIC-Lasso, show similar performance, but they exhibit high variance, indicating inconsistency in their ability to provide reliable explanations. Looking at concrete examples reveals that all methods are able to identify high-impact explanation triples. However, KGEPrisma with HSCI-Lasso is the most consistent over the 10 runs in identifying such high-impact explanation triples. Thus, similar to the results for DistMult, KGEPrisma with HSIC-Lasso consistently generates more sensible explanations for ConvE on FB15k-237.

The overall strong performance of KGEPrisma on FB15k-237 across all models can be attributed to its capacity to reconstruct the decision surface of the KGE model locally. This allows it to produce explanation subgraphs that are highly aligned with the model's decision-making process.
A similar effect can be observed on WN18RR. 

\begin{table*}[ht]
\centering
\caption{Results for WN18RR. The results are the mean and variance of the MRR and Hits@1 over ten runs; the lower the MRR and Hits@1, the better. The best results are bold.}
\resizebox{\textwidth}{!}{%
\begin{tabular}{ c | c c | c c | c c }
 & \multicolumn{2}{c | }{TransE} & \multicolumn{2}{c|}{DistMult} & \multicolumn{2}{c}{ConvE} \\
\hline
 & MRR & Hit@1 & MRR & Hit@1 & MRR & Hit@1 \\
\hline
Retraining & 0.86 $\pm$ 0.04 & 0.78 $\pm$ 0.07 & 0.92 $\pm$ 0.08 & 0.88 $\pm$ 0.12 & 0.74 $\pm$ 0.28 & 0.62 $\pm$ 0.31 \\
\hline
Global Random & 0.75 $\pm$ 0.20 & 0.61 $\pm$ 0.27 & 0.91 $\pm$ 0.04 & 0.86 $\pm$ 0.06 & 0.65 $\pm$ 0.25 & 0.57 $\pm$ 0.23 \\
Local Random & 0.76 $\pm$ 0.22 & 0.65 $\pm$ 0.28 & 0.92 $\pm$ 0.03 & 0.87 $\pm$ 0.05 & 0.67 $\pm$ 0.28 & 0.58 $\pm$ 0.28 \\
\hline
AnyBURLExplainer & 0.26 $\pm$ 0.06 & 0.16 $\pm$ 0.06 & \textbf{0.22} $\pm$ 0.05 & \textbf{0.14} $\pm$ 0.04 & 0.17 $\pm$ 0.13 & 0.02 $\pm$ 0.07 \\
Data Poisoning & 0.42 $\pm$ 0.20 & 0.32 $\pm$ 0.23 & 0.39 $\pm$ 0.05 & 0.33 $\pm$ 0.06 & 0.07 $\pm$ 0.05 & \textbf{0.00} $\pm$ 0.00 \\
Kelpie & 0.37 $\pm$ 0.11 & 0.28 $\pm$ 0.11 & 0.23 $\pm$ 0.05 & 0.16 $\pm$ 0.05 & 0.08 $\pm$ 0.05 & \textbf{0.00} $\pm$ 0.00 \\
\hline
KGEPrisma - K-Lasso & \textbf{0.05} $\pm$ 0.05 & \textbf{0.02} $\pm$ 0.06 & 0.37 $\pm$ 0.07 & 0.30 $\pm$ 0.08 & 0.06 $\pm$ 0.11 & \textbf{0.00} $\pm$ 0.00 \\
KGEPrisma - HSIC-Lasso & 0.10 $\pm$ 0.05 & 0.04 $\pm$ 0.04 & 0.38 $\pm$ 0.04 & 0.31 $\pm$ 0.04 & \textbf{0.01} $\pm$ 0.01 & \textbf{0.00} $\pm$ 0.00 \\
KGEPrisma - MDI & 0.15 $\pm$ 0.10 & 0.10 $\pm$ 0.11 & 0.26 $\pm$ 0.06 & 0.20 $\pm$ 0.05 & \textbf{0.01} $\pm$ 0.01 & \textbf{0.00} $\pm$ 0.00 \\
\end{tabular}
}
\label{tab:results_wn18rr}
\end{table*}

The results for \textbf{WN18RR} in Table~\ref{tab:results_wn18rr} also indicate state-of-the-art performance of KGEPrisma. This further emphasizes the importance of localizing explanations around the triples that the model identifies as similar (see Figure~\ref{fig:overview}, Step 1). 

For the TransE model, KGEPrisma outperforms the other XAI methods across all surrogates, with K-Lasso being the best-performing surrogate. However, the variance indicates that the performance differences among the three surrogates are minimal. 

Examining the explanations generated by KGEPrisma and the second-best performing method, AnyBURLExplainer, reveals that both can generally identify the same explanations. For instance, they consistently explain the predicted $hypernym$ relationship between the entities $Platichthys$ and $fish\_genus$ using its inverse, the $member\_meronym$ relationship. Nonetheless, there are cases where the inverse $member\_meronym$ relation does not exist to reconstruct the $hypernym$ relationship due to the incompleteness of the knowledge graph. In such instances, AnyBURLExplainer fails to find an explanation, while KGEPrisma can construct more extensive and complex explanation chains, as can be observed with the $hypernym$ relationship between the entity $family\_Treponemataceae$ and $bacteria\_family$. The correct explanation triples look as follows:
\begin{equation*}
\scalebox{0.7}{$
\begin{aligned}
(order\_Spirochaetales,\ member\_meronym,\ family\_Treponemataceae), \\ 
(division\_Eubacteria,\ member\_meronym,\ order\_Spirochaetales), \\
(division\_Eubacteria,\ member\_meronym,\ bacteria\_family)
\end{aligned}
$}
\end{equation*}

This explanation shows that the Treponemataceae family is a meronym of the Spirochaetales order, and the Spirochaetales order is a member of the Eubacteria division which in turn also has the bacteria family as a member. As a result, the bacteria family is also a hypernym of Treponemataceae. KGEPrisma successfully identified this complex relationship due to its localized approach to explanation, while AnyBURLExplainer was unable to do so.

\begin{figure*}[!ht]
\centering
\includegraphics[width=0.8\textwidth]{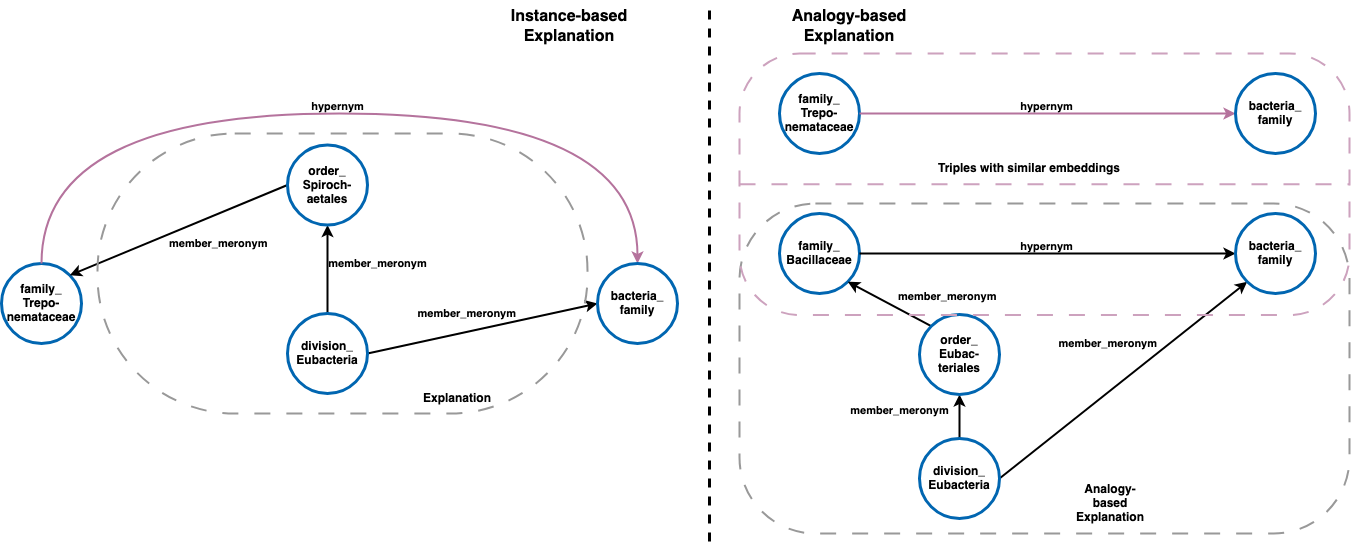}
\caption{
TransE in WN18RR predicts that the bacteria family serves as a hypernym for the Treponemataceae family. An instance-based explanation provided by KGEPrisma explains this relationship by indicating that the Treponemataceae family is a subset (or meronym) of the Spirochaetales order. This order belongs to the Eubacteria division, which also includes the the bacteria family. Thus, the instance-based explanation effectively demonstrates why the bacteria family can be considered a hypernym for the Treponemataceae family.
Supporting the instance-based explanation, the analogy-based explanation provided by KGEPrisma points out to the user that the triple $(family\_Bacillaceae, hypernym, bacteria\_family)$ is similar to the predicted triple. In the context of this triple, we know that the Bacillaceae family is a subset of the Eubacteriales order, which in turn belongs to the Eubacteria division that includes the bacteria family. This analogous example reinsures the user that the reasoning pattern leading to the prediction aligns with other established facts within the KG.}
\label{fig:bacteria}
\end{figure*}

For DistMult on WN18RR, AnyBURLExplainer performs the best. However, due to the variance in results, it is comparable to KGEPrisma using the MDI surrogate and Kelpie. When examining actual explanation examples, AnyBURLExplainer, KGEPrisma and Kelpie tend to produce similar explanations.

For ConvE on WN18RR, a similar trend can be observed. KGEPrisma shows the best performance, although it only marginally outperforms the other methods. This slight advantage is primarily due to KGEPrisma being more consistent in identifying the relevant explanation triples, as indicated by the variance and validated by the example explanations.

Once again, KGEPrisma outperforms AnyBURLExplainer, Kelpie, and Data Poisoning due to its localization around the predicted triple, consistently leading to accurate explanations. This observation also holds true when looking at the last benchmark knowledge graph, Kinship.

\begin{table*}[ht]
\centering
\caption{Results for Kinship. The results are the mean and variance of the MRR and Hits@1 over ten runs; the lower the MRR and Hits@1, the better. The best results are bold.}
\resizebox{\textwidth}{!}{%
\begin{tabular}{ c | c c | c c | c c }
 & \multicolumn{2}{c | }{TransE} & \multicolumn{2}{c|}{DistMult} & \multicolumn{2}{c}{ConvE} \\
\hline
 & MRR & Hit@1 & MRR & Hit@1 & MRR & Hit@1 \\
\hline
Retraining & 0.89 $\pm$ 0.08 & 0.75 $\pm$ 0.09 & 0.64 $\pm$ 0.13 & 0.55 $\pm$ 0.07 & 0.94 $\pm$ 0.05 & 0.92 $\pm$ 0.04 \\
\hline
Global Random & 0.83 $\pm$ 0.03 & 0.77 $\pm$ 0.05 & 0.57 $\pm$ 0.12 & 0.49 $\pm$ 0.08 & 0.86 $\pm$ 0.07 & 0.78 $\pm$ 0.10 \\
Local Random & 0.80 $\pm$ 0.06 & 0.73 $\pm$ 0.09 & 0.53 $\pm$ 0.13 & 0.045 $\pm$ 0.03 & 0.83 $\pm$ 0.05 & 0.75 $\pm$ 0.08 \\
\hline
AnyBURLExplainer & 0.60 $\pm$ 0.07 & 0.33 $\pm$ 0.12 & \textbf{0.04} $\pm$ 0.02 & \textbf{0.00} $\pm$ 0.01 & 0.72 $\pm$ 0.06 & 0.57 $\pm$ 0.08 \\
Data Poisoning & \textbf{0.05} $\pm$ 0.03 & \textbf{0.01} $\pm$ 0.04 & 0.10 $\pm$ 0.04 & 0.04 $\pm$ 0.04 & 0.75 $\pm$ 0.04 & 0.61 $\pm$ 0.06 \\
Kelpie & \textbf{0.05} $\pm$ 0.03 & \textbf{0.01} $\pm$ 0.02 & 0.10 $\pm$ 0.06 & 0.03 $\pm$ 0.05 & 0.72 $\pm$ 0.05 & 0.56 $\pm$ 0.08 \\
\hline
KGEPrisma - K-Lasso & 0.52 $\pm$ 0.07 & 0.28 $\pm$ 0.09 & 0.05 $\pm$ 0.03 & \textbf{0.00} $\pm$ 0.00 & \textbf{0.65} $\pm$ 0.09 & \textbf{0.50} $\pm$ 0.12 \\
KGEPrisma - HSIC-Lasso & 0.58 $\pm$ 0.07 & 0.33 $\pm$ 0.11 & \textbf{0.04} $\pm$ 0.01 & \textbf{0.00} $\pm$ 0.00 & 0.72 $\pm$ 0.07 & 0.57 $\pm$ 0.10 \\
KGEPrisma - MDI & 0.51 $\pm$ 0.06 & 0.23 $\pm$ 0.07 & 0.08 $\pm$ 0.04 & 0.03 $\pm$ 0.05 & 0.70 $\pm$ 0.08 & 0.55 $\pm$ 0.10 \\
\end{tabular}
}
\label{tab:results_kinship}
\end{table*}

In the \textbf{Kinship} KG (cf. Table~\ref{tab:results_kinship}), KGEPrisma performs best for two of the three KGE models.
For the TransE model, Kelpie, as well as Data Poisoning, outperforms KGEPrisma and AnyBURLExplainer. Looking at example predictions and explanations reveals why. The adversarial methods Kelpie and Data Poisoning robustly identify existing inverse relations, as what TransE relies on the most to reconstruct missing links. For example, the reconstruction of the relation $term2$ works consistenly with $term1$ as its inverse. The other XAI Methods have a higher chance of failing to identify such inverse relations, meanwhile coming up with different patterns which are less faithful to the model behaviour while still giving a sensible justification for why the link was predicted. This can also be observed in the still acceptable performance of KGEPrisma and AnyBURLExplainer in terms of faithfulness. 
For DistMult, all models perform comparatively well. KGEPrisma and AnyBURLExplainer share place one. KGEPrisma with the HSIC-Lasso surrogate, though, exhibits a slightly lower variance. The example predictions and explanations also show that they mostly come up with the same explanations. For example, for the prediction $(person23,\ term15,\ person29)$, both find the explanation $(person29,\ term5,\ person23)$.
The faithfulness results for ConvE show that it is harder to explain its model behaviour. All models outperform the retraining and random baselines. However, after removing the explanations from the training graph, ConvE is still, even in the case of the most faithful explanations by KGEPrisma, able to reconstruct the predictions with a Hit@1 of 50\%. Looking at the example predictions and explanations gives no visual clue as to why this is the case. The explanations of triples look comparable to the ones observed for the other KGE models.

Overall, KGEPrisma performs strongly for all models and KG's, demonstrating its ability to explain a wide variety of KGE models robustly.

\subsection{Runtime Evaluation} 

\begin{figure*}[htbp]

  \centering
  \begin{minipage}{0.32\textwidth}
    \centering
   \begin{tikzpicture}[scale=0.7]
\begin{axis}[
    title={TransE},
    ybar,
    bar width=15pt,
    enlarge x limits=0.2,
    symbolic x coords={
        AnyBURLExplainer,
        Kelpie,
        Data\ Poisoning,
        KGEPrisma\ MDI,
        KGEPrisma\ HSIC-Lasso,
        KGEPrisma\ K-Lasso
    },
    xtick=data,
    xticklabel style={rotate=45, anchor=east},
    ylabel={Time (s)},
    nodes near coords,
    ymin=0,
    grid=both,
    major grid style={line width=.2pt, draw=gray!50},
    minor grid style={line width=.1pt, draw=gray!20},
    axis background/.style={fill=none}, 
]
\addplot[fill=teal!60, draw=gray] coordinates { 
    (AnyBURLExplainer,104.81444)
    (Kelpie,1879.17318)
    (Data\ Poisoning,5.60188)
    (KGEPrisma\ MDI,8.01996)
    (KGEPrisma\ HSIC-Lasso,8.79303)
    (KGEPrisma\ K-Lasso,7.17646)
};
\end{axis}
\end{tikzpicture}
\label{fig:time_kinship_transe}
  \end{minipage}
  \hfill
  \begin{minipage}{0.32\textwidth}
    \centering
   \begin{tikzpicture}[scale=0.7]
\begin{axis}[
    title={DistMult},
    ybar,
    bar width=15pt,
    enlarge x limits=0.2,
    symbolic x coords={
        AnyBURLExplainer,
        Kelpie,
        Data\ Poisoning,
        KGEPrisma\ MDI,
        KGEPrisma\ HSIC-Lasso,
        KGEPrisma\ K-Lasso
    },
    xtick=data,
    xticklabel style={rotate=45, anchor=east},
    ylabel={Time (s)},
    nodes near coords,
    ymin=0,
    grid=both,
    major grid style={line width=.2pt, draw=gray!50},
    minor grid style={line width=.1pt, draw=gray!20},
    axis background/.style={fill=none}, 
]
\addplot[fill=teal!60, draw=gray] coordinates { 
    (AnyBURLExplainer,105.49226)
    (Kelpie,724.41485)
    (Data\ Poisoning,4.75755)
    (KGEPrisma\ MDI,5.81739)
    (KGEPrisma\ HSIC-Lasso,7.14538)
    (KGEPrisma\ K-Lasso,5.09598)
};
\end{axis}
\end{tikzpicture}
\label{fig:time_kinship_distmult}

\end{minipage}
  \hfill
  \begin{minipage}{0.32\textwidth}
    \centering
   \begin{tikzpicture}[scale=0.7]
\begin{axis}[
    title={ConvE},
    ybar,
    bar width=15pt,
    enlarge x limits=0.2,
    symbolic x coords={
        AnyBURLExplainer,
        Kelpie,
        Data\ Poisoning,
        KGEPrisma\ MDI,
        KGEPrisma\ HSIC-Lasso,
        KGEPrisma\ K-Lasso
    },
    xtick=data,
    xticklabel style={rotate=45, anchor=east},
    ylabel={Time (s)},
    nodes near coords,
    ymin=0,
    grid=both,
    major grid style={line width=.2pt, draw=gray!50},
    minor grid style={line width=.1pt, draw=gray!20},
    axis background/.style={fill=none}, 
]
\addplot[fill=teal!60, draw=gray] coordinates { 
    (AnyBURLExplainer,106.18053)
    (Kelpie,520.09450)
    (Data\ Poisoning,8.07398)
    (KGEPrisma\ MDI,11.38097)
    (KGEPrisma\ HSIC-Lasso,11.59298)
    (KGEPrisma\ K-Lasso,9.71097)
};
\end{axis}
\end{tikzpicture}
\label{fig:time_kinship_conve}
\end{minipage}
  \caption{Mean execution time (over 10 runs on the same hardware, 40 explanations per run) for AnyBURLExplainer, Kelpie, Data Poisoning, and KGEPrisma explaining TransE, DistMult, and ConvE in the Kinship KG.}
  \label{fig:time_kinship_all}
\end{figure*}

KGEPrisma also performs well, runtime-wise, compared to the other XAI methods.

This can be observed, for example, in the benchmark knowledge graph Kinship. 
Figure~\ref{fig:time_kinship_all} shows the mean runtime in seconds over ten runs of the XAI methods AnyBURLExplainer, Kelpie, Data Poisoning and KGEPrisma, explaining TransE, DistMult and ConvE in Kinship. 

The setup and implementation used is the same as in the faithfulness evaluation. The runtime experiment is executed on an AWS EC2 g5.12xlarge \footnote{Link to EC2 instance description: https://aws.amazon.com/de/ec2/instance-types/g5}.

The results show that KGEPrisma calculates explanations in Kinship for all KGE models by a factor of approximately 10 to 20 times faster compared to AnyBURLExplainer and approximately 50 to 200 times faster compared to Kelpie.
On the one hand, this performance gain is due to KGEPrisma's efficient localisation approach, approximating the KGE model's local decision surface by looking at the subgraphs sampled around triples the model learned to see as similar as described in Subsections~\ref{subsec:knn-search} to \ref{subsec:surrogate_models} (cf. Figure~\ref{fig:overview} Step 1-3). Approximating the local decision surface results in a compact learning space for the surrogate model and, thus, a quick run-time. Kelpie, on the other hand, relies on expensive perturbation over triples to calculate an explanation. The perturbation is guided by a heuristic to focus on likely relevant triples. However, this process is still runtime-wise expensive. 
AnyBURLExplainer has a constant run-time for all KGE models, as it does not depend on the underlying KGE model to calculate its explanations. Its execution time is set by the authors to a constant, meaning that it will return the best explanation found within the set timeframe \cite{ijcai2022p391}.  

Overall, the results demonstrate that KGEPrisma is a time-efficient method to post-hoc explain KGE models while delivering state-of-the-art faithfulness values.

\subsection{Qualitative Discussion of the Three Explanation Modalities}

The previous evaluation demonstrated that KGEPrisma performs at a state-of-the-art level in faithfulness while maintaining efficient runtime. Additionally, KGEPrisma is capable of generating three distinct types of explanations, rule-based, analogy-based, and instance-based, which present multiple perspectives on a single predicted triple to a user. The purpose of these different explanation types is to explain a single prediction from various perspectives to the user. This approach aims to ensure a better-informed user and, ultimately, to foster greater trust in the predicted link. The following qualitative evidence illustrates the effectiveness of these explanation modalities from the user's perspective.

Consider, for example, the prediction from the WN18RR KG that classifies the bacteria family as a hypernym of the Treponemataceae family. The instance-based explanation (cf. Figure~\ref{fig:bacteria}) for this prediction is provided in the form of a series of triples: 
\begin{equation*}
\scalebox{0.7}{$
\begin{aligned}
(order\_Spirochaetales,\ member\_meronym,\ family\_Treponemataceae), \\ 
(division\_Eubacteria,\ member\_meronym,\ order\_Spirochaetales), \\
(division\_Eubacteria,\ member\_meronym,\ bacteria\_family).
\end{aligned}
$}
\end{equation*} 
This explanation demonstrates that the Treponemataceae family is a subset (i.e., a meronym) of the Spirochaetales order, which in turn belongs to the Eubacteria division, a group that also includes the broader category referred to as the bacteria family. In this way, the instance-based explanation clarifies why the bacteria family can be seen as a hypernym of the Treponemataceae family.

While the instance-based modality provides focused context relevant to the specific prediction, it is limited by the narrow scope of the predicted triple itself. To capture a more general pattern within the domain, the user can look at the rule-based explanation. For example, the rule-based explanations for the same predicted triple states states: 
\begin{equation*}
\scalebox{0.7}{$
\begin{aligned}
hypernym(family\_Treponemataceae, bacteria\_family) \  \leftarrow \\
member\_meronym(X \in noun.animal, family\_Treponemataceae) \  \land \\ member\_meronym(Y \in noun.animal, X) \ \land \\ member\_meronym(Y, bacteria\_family).
\end{aligned}
$}
\end{equation*}
The rule states that the Treponemataceae family is hypernymic to the bacteria family, because it appears in a recurrent pattern that connects both entities, as evidenced by multiple instances across the knowledge graph. Such a rule indicates that the reasoning behind the prediction is not a one-off occurrence but rather reflects a recurring pattern in the domain, thereby reinforcing user confidence.

In addition, an analogy-based explanation (cf. Figure~\ref{fig:bacteria}) provides further support by comparing the prediction with another known instance. The knowledge graph embedding model learned in WN18RR that ${family\_Bacillaceae}$ is closely positioned to $family\_Treponemataceae$ in the latent space. The analogous explanation then outlines a similar set of relationships: 
\begin{equation*}
\scalebox{0.7}{$
\begin{aligned}
(order\_Eubacteriales, member meronym, family\_Bacillaceae), \\
(division Eubacteria, member meronym, order\_Eubacteriales), \\
(division Eubacteria, member meronym, bacteria\_family).
\end{aligned}
$}
\end{equation*} 
This analogous case reaffirms that the reasoning pattern resulting in the prediction is consistent with other established facts within the knowledge graph.

Collectively, these explanation modalities, each providing a distinct vantage point, contribute to a better understanding of the predictions by the user, leading to more trust in the model's predictions.

While achieving state-of-the-art results, KGEPrisma has several limitations.

\subsection{Limitations}

Although KGEPrisma is compatible with most KGE models, the method requires the KGE model to operate in Euclidean space, as the distance function used in Step 1 (cf. Section~\ref{subsec:knn-search}) is defined explicitly for such spaces. Consequently, it cannot post-hoc explain models like MuRe \cite{balavzevic2019multi} and QuatE \cite{zhang2019quaternion}, which utilize non-Euclidean spaces.

Additionally, the choice of the hyperparameter $k$ that determines the number of nearest neighbors in Step 1 (cf. Section~\ref{subsec:knn-search}) influences the quality of the explanation. If set too high, it can lead to explanations that are not localized around the instance to be explained and include noise, reducing their accuracy and relevance. The ideal value depends on the KGE model, as explained by KGEPrisma. Ideally, $k$ should be set large enough to encompass all embeddings within a cluster. The elbow method can be employed to determine the ideal value \cite{Thorndike_1953}.

The runtime of KGEPrisma is sensitive to the node degree of the KG. Higher node degrees result in an increased number of paths for expansion in Step 3 (cf. Section~\ref{subsec:mining_clauses}), which significantly extends the runtime for nodes with high connectivity.

Lastly, KGEPrisma assumes that KGE models learn embeddings solely from the symbolic structure of the knowledge graph. This limits its applicability to models incorporating literals, such as textual data, leading to less faithful explanations for such KGE models. 

\section{Related Work}

Explainable Artificial Intelligence is about mapping the input of a black-box model to its output. That way, XAI methods compute an explanation of the model’s behaviour. The explanation is supposed to justify the model's output \cite{adadi2018peeking,lipton2018mythos}, helping to identify risks and flaws in the black-box model. 

A common approach to achieve this are attribution methods. Attribution methods assign a value to input features resembling how relevant they are to the output. Among the notable local post-hoc XAI methods is LIME \cite{ribeiro2016lime}. It utilizes local surrogate models to approximate the behavior of black-box models around specific instances, thereby providing local interpretability. Extending this concept, SHAP \cite{lundberg2017shap} employs Shapley values to calculate attributions across all possible coalitions of features. Integrated Gradients \cite{sundararajan2017ig} calculates the path integral of gradients along the straight line from a baseline to the input, highlighting the contribution of each feature to the difference in output. LRP \cite{Montavon2019lrp} backpropagates the output to the input layer, redistributing relevance scores across layers to identify relevant features. DEEPLift \cite{shrikumar2017deeplift} compares activation's to a reference activation, allocating relevance scores based on the difference, thus observing the shift caused by each input feature. These methods effectively assign the contributions of input features in standard settings such as tabular, image, or textual data. However, they are not trivial to apply to KGE models due to the complex and latent nature of the input triple, where simple attribution to input dimensions offers minimal insights into the predictive mechanisms.

Several XAI methods have been adapted for KGE models that employ graph neural networks (GNNs), yet their application remains constrained by the specific architectures and mechanisms of GNNs. GraphLIME \cite{huang2023graphlime} leverages local perturbation and the HSIC-Lasso as a surrogate model to approximate decision surfaces within GNNs. PGMExplainer \cite{vu2020pgmx} generates a probabilistic graphical model that captures the Markov blanket of the target prediction, though it is computationally expensive due to its reliance on input perturbation. GraphLRP \cite{schnake2022graphlrp} adapts LRP for GNNs by propagating relevance through their aggregation functions. These methods, designed around the unique operations of GNN-based KGEs, are not universally applicable across the broader spectrum of KGE models, which often do not use GNN-based frameworks.

For that reason, inherently interpretable KGE models were introduced.
DistMult \cite{yang2015distmult}, for example, introduces a unified framework for learning entity and relation representations using neural embeddings, emphasizing the extraction of logical rules from relation embeddings, which is primarily focused on capturing relational semantics through matrix operations. IterE \cite{wen2019itere} extends this by iteratively learning embeddings and rules to complement the weaknesses of each method, particularly enhancing the embeddings of sparse entities through rule incorporation. ExCut \cite{Gad-Elrad2020} similarly combines embeddings with rule mining but shifts its focus towards generating interpretable entity clusters and iteratively refining them with rules derived from embedding patterns. However, these approaches are tailored to one specific KGE model and do not generalise to other models.

One stream of work that is applicable to KGE models comes from research on adversarial attacks. 
The work of \cite{pezeshkpour2019investigating} introduces gradient-based adversarial attacks, emphasizing the identification of influential training facts to test model sensitivity and robustness. 
Similarly, Data Poisoning \cite{zhang2019data} targets the robustness of KGEs to adversarial attacks by proposing strategies for data poisoning that directly manipulate the knowledge graph.
The approach by \cite{bhardwaj-etal-2021-adversarial} uses model-agnostic instance attribution methods from interpretable machine learning to select adversarial deletions, focusing on data poisoning to influence KGE predictions.
This contrasts with AnyBURLExplainer \cite{ijcai2022p391}. Which uses rule learning and abductive reasoning to perform adversarial attacks independent of the model’s internal workings, thus focusing on explanation via adversarial attacks. AnyBURLExplainer outperformed other adversarial methods in its evaluation study. 
KGEPrisma deviates from these methods by focusing not on crafting adversarial inputs to disrupt the model but on decoding and interpreting the latent representations created by KGE models.

Other work focuses on explaining KGE models with surrogate models. 
The work by \cite{ruschel2024xke,gusmao2018interpreting} and \cite{polleti2019faithfully} involves surrogate models that attempt to decode the embeddings through global and local perspectives respectively, with the former utilizing context-aware heuristics and the latter focusing on neighborhood features without capturing multi-hop dependencies. Meanwhile, KGEx \cite{baltatzis2023kgex} leverages multiple training subsets to generate surrogate models that provide explanations based on training data impact. The approach by \cite{islam2022negative} incorporates rule mining with embedding learning, independent of the KGE model.
Yet other work such as OxKBC \cite{nandwani2020oxkbc} and CPM \cite{stadelmaier2019cpm} generate human-understandable explanations through heuristic templates and context paths, respectively, but struggle with scalability and faithfulness to underlying KGE models. KE-X \cite{zhao2023kex} leverages information entropy for subgraph analysis, improving interpretability but lacking clear heuristics for reconstructing model behavior. FeaBI \cite{ismaeil2023feabi} constructs interpretable vectors for entity embeddings to provide model explanations, while \cite{chandrahas2020inducing} focuses on semantic interpretability through entity co-occurrence statistics. Additionally, Kelpie \cite{rossi2022kelpie} and KGExplainer \cite{kgexplainer_ma} introduce perturbation-based frameworks requiring retraining, which is resource-intensive.
In contrast to these methods, KGEPrisma decodes the latent representations of KGE models. It identifies symbolic regularities in the subgraph neighborhood of the predicted link to generate rule-based, instance-based and analogy-based explanations. This approach remains faithful to the model’s behavior and is computationally inexpensive, as it does not depend on perturbing training data or retraining the model. 
\section{Conclusion}

This paper presented KGEPrisma, a novel post-hoc explainable AI method explicitly designed for KGE models. Despite their utility in knowledge graph completion, KGE models face criticism due to their black-box nature. KGEPrisma directly decodes these models' latent representations by identifying symbolic patterns within the subgraph neighborhoods of entities with similar embeddings. By translating these patterns into human-readable rules and facts, the method provides clear, interpretable explanations that bridge the gap between the abstract representations and predictive outputs of KGE models.
This work contributes a post-hoc and local explainable AI approach that requires no retraining of the KGE model, is faithful to model predictions and can adapt to various explanation styles (rule-based, instance-based and analogy-based). Extensive evaluations demonstrated that this method outperforms state-of-the-art approaches, offering a distinct advantage by remaining faithful to the underlying predictive mechanisms of KGE models.
Future research will apply KGEPrisma to knowledge graph domains, such as the biomedical field, where explainability is critical. Here, clear and interpretable results can improve decision-making and foster trust in AI-based predictions. By providing transparent insights into the patterns and rules guiding KGE models, KGEPrisma uncovers hidden decision-making patterns in KGE models. Let us make KGE models understandable and trustworthy in high-risk use cases.

\section*{Acknowledgments}
We thank Sony AI for funding the research and Ute Schmid for enabling the collaboration.

\appendix
\section{Triple Embedding Extraction} \label{app:triple_embeddings}

The construction of triple embeddings depends on the specific interaction function $i$ of the Knowledge Graph Embedding model. Generally, triple embeddings represent the learned interaction between the head entity ($e^{head}$), relation ($r$), and tail entity ($e^{tail}$) before aggregation into a scalar score. This stage retains the richest representation of the entity dynamics. Below, the triple embeddings $v_{triple}$ for the three evaluated KGE models are described in detail.

\subsection{TransE}
\label{app:TransE}
The interaction function for TransE is:

\begin{equation}
    i(e^{head}, r, e^{tail}) = -\| v^{head} + v^r - v^{tail} \|_2
\end{equation}

where $v^{head}, v^r, v^{tail} \in \mathbb{R}^n$ are the learned embeddings of $e^{head}$, $r$, and $e^{tail}$, respectively \cite{bordes2013translating}.
Since $i(e^{head},r,t)$ contains no additional parameters, it does not introduce information beyond the input embeddings. The triple embedding, denoted as $v_{triple}$, consolidates the learned embeddings of $e^{head}$, $r$, and $e^{tail}$ before scoring:

\begin{equation}
v_{triple}=[v^{head};v^r;v^{tail}],
\end{equation}

where $[;]$ denotes vector concatenation.

\subsection{DistMult}
\label{app:DistMult}
The interaction function for DistMult is:

\begin{equation}
    i(e^{head}, r, e^{tail}) = \sum_{i=1}^n v^{head}_i v^r_i v^{tail}_i,
\end{equation}

where $v^{head}, v^r, v^{tail} \in \mathbb{R}^n$ are the learned embeddings of $e^{head}$, $r$, and $e^{tail}$, respectively \cite{yang2015distmult}.
As with TransE, this function has no learned parameters, merely aggregating $v^{head}$, $v^r$ and $v^{tail}$ into a score. The triple embedding for DistMult is similarly defined as:

\begin{equation}
    v_{triple}=[v^{head};v^r;v^{tail}].
\end{equation}

\subsection{ConvE}
\label{app:ConvE}
ConvE differs from TransE and DistMult as its interaction function incorporates learned parameters, enriching the information within the triple embedding \cite{dettmers2018conve}.
For input embeddings $v^{head},v^r,v^{tail} \in \mathbb{R}^d$, ConvE first combines $v^{head}$ and $v^r$ into a matrix $\mathbf{A} \in \mathbb{R}^{2 \times d}$, where the rows represent $v^{head}$ and $v^r$, respectively. $\mathbf{A}$ is reshaped into $\mathbf{B} \in \mathbb{R}^{m \times n}$, splitting its rows to represent $v^{head}$ and $v^r$. A set of 2D convolutional filters $\Omega = \{\mathbf{\omega}_i \mid \mathbf{\omega}_i \in \mathbb{R}^{r \times c}\}$ is applied to $\mathbf{B}$ to capture interactions between $v^{head}$ and $v^r$.
The resulting feature maps are reshaped and concatenated into a feature vector $\mathbf{v} \in \mathbb{R}^{|\Omega| \cdot r_c}$, which is then mapped into the entity space via a linear transformation:

\begin{equation}
    v^{head,r} = \mathbf{v}^\top \mathbf{W},
\end{equation}

where $\mathbf{W} \in \mathbb{R}^{|\Omega| \cdot r_c \times d}$. Finally, the interaction function aggregates the enriched representation with $v^{tail}$ through:

\begin{equation}
    i(e^{head}, r, e^{tail}) = v^{head,r} v^{tail}.
\end{equation}

The enriched representation $v^{head,r}$ combines and processes $v^{head}$ and $v^r$ through convolution, normalization, and dense layers, capturing more information than the individual embeddings \cite{ali2020benchmarking}. Thus, the triple embedding for ConvE concatinates $v^{head,r}$ and $v^{tail}$:

\begin{equation}
    v_{triple}=[v^{head,r};v^{tail}].
\end{equation}


\bibliographystyle{kr}
\bibliography{cite}

\end{document}